# The OpenLAM Challenges

Anyang Peng[1], Xinzijian Liu[1,2], Ming-Yu Guo[2,5], Linfeng Zhang[1,2,*] and Han Wang[3,4,*]

[1] AI for Science Institute, Beijing 100080, P. R. China
[2] DP Technology, Beijing 100080, P. R. China
[3] HEDPS, CAPT, College of Engineering, Peking University, Beijing 100871, P.R. China
[4] Laboratory of Computational Physics, Institute of Applied Physics and Computational Mathematics, Fenghao East Road 2, Beijing 100094, P.R. China
[5] School of Chemistry, Sun Yat-sen University, Guangzhou 510006, P.R. China

E-mail: linfeng.zhang.zlf@gmail.com; wang_han@iapcm.ac.cn

**Abstract**

Inspired by the success of Large Language Models (LLMs), the development of Large Atom Models (LAMs) has gained significant momentum in scientific computation. Since 2022, the Deep Potential team has been actively pretraining LAMs and launched the OpenLAM Initiative to develop an open-source foundation model spanning the periodic table. A core objective is establishing comprehensive benchmarks for reliable LAM evaluation, addressing limitations in existing datasets. As a first step, the LAM Crystal Philately competition has collected over 19.8 million valid structures, including 1 million on the OpenLAM convex hull, driving advancements in generative modeling and materials science applications.

**1. Overview**

Over the past decade, the field of natural language processing (NLP) has witnessed remarkable advancements. The widespread success and ubiquity of Large Language Models (LLMs) or foundational models can be attributed to three pivotal factors: (1) the introduction of the transformer architecture, which significantly enhances model capabilities; (2) access to an enormous corpus of linguistic data encompassing almost every conceivable topic; and (3) the extraordinary generalizability enabled by advanced pretraining techniques. These cornerstones have been instrumental in driving the development of powerful AI tools like ChatGPT. Drawing a parallel to scientific computation, we envisions the emergence of a general Large Atom Model (LAM). Achieving such a breakthrough will similarly require the establishment of these foundational elements.

We have been actively pretraining Large Atom Models (LAMs) since 2022 [1][2]. Over this time, we have observed growing interest in this field from both academia and industry [3-7]. We believe that open-source and transparency will play an vital role in the advancement of LAMs. To foster collaboration and innovation, we are thrilled to launch the OpenLAM Initiative [8]—a community-driven effort to develop an open-source foundation model that spans the entire periodic table.

A crucial objective of the OpenLAM Initiative is to develop a comprehensive benchmark for the reliable and confident evaluation of LAMs. While some existing benchmarks or evaluation datasets focus on specific tasks [9], or are limited to a restricted chemical space [10][11], they fall short of capturing the full capacity of LAMs. Recent efforts [12] have sought to evaluate foundation models in realistic application contexts; however, we believe that a community-driven initiative is essential to create a comprehensive benchmark that can serve as a consensus standard for the field.

A benchmark set fully reflecting the potential of LAMs should span over an extensive configurational and chemical space. Thus, the LAM Crystal Philately competition is launched as an initial step in this direction.

## 2. Challenge Introduction

The LAM Crystal Philately competition [13] aims to collect stable crystal structures that can expand the configurational and chemical space accessible to Large Atom Models (LAMs). Participants are encouraged to use any generative algorithm and any database to construct crystal structures that meet the competition's requirements. This effort will enable the creation of an open-source database that serves as both a reservoir for building task-specific benchmarks and a resource for enriching LAM training data. Additionally, this database plays a crucial role in providing structures for new material discovery. In the future, for computational-aided materials design studies, researchers will tend not to create structures from scratch, but filter materials from existing databases. Thus, our database from this competition provides an essential resource for such endeavors.

In the LAM Crystal Philately competition, participants will submit crystal structures in the form of crystallographic information files (CIFs). Each submission will undergo evaluation through the LAM Crystal Philately workflow, which consists of three phases: (1) Pre-submission Phase: The validated CIF submissions are parsed into crystal structures, and then undergo a full geometry relaxation using a Machine Learning Interatomic Potential (MLIP) model; (2) Sampling Phase: A quality check will run over a randomly selected subset of each submission. Only those with at least 30% passing rate can be viewed as a successful submission and enter the next Phase; (3) Evaluation Phase: In this phase, all structures undergo a process similar to the sampling phase. Each valid structure is scored based on its energy above the hull, as calculated by the MLIP model, and is added to the competition's open-source database. (Figure 1)

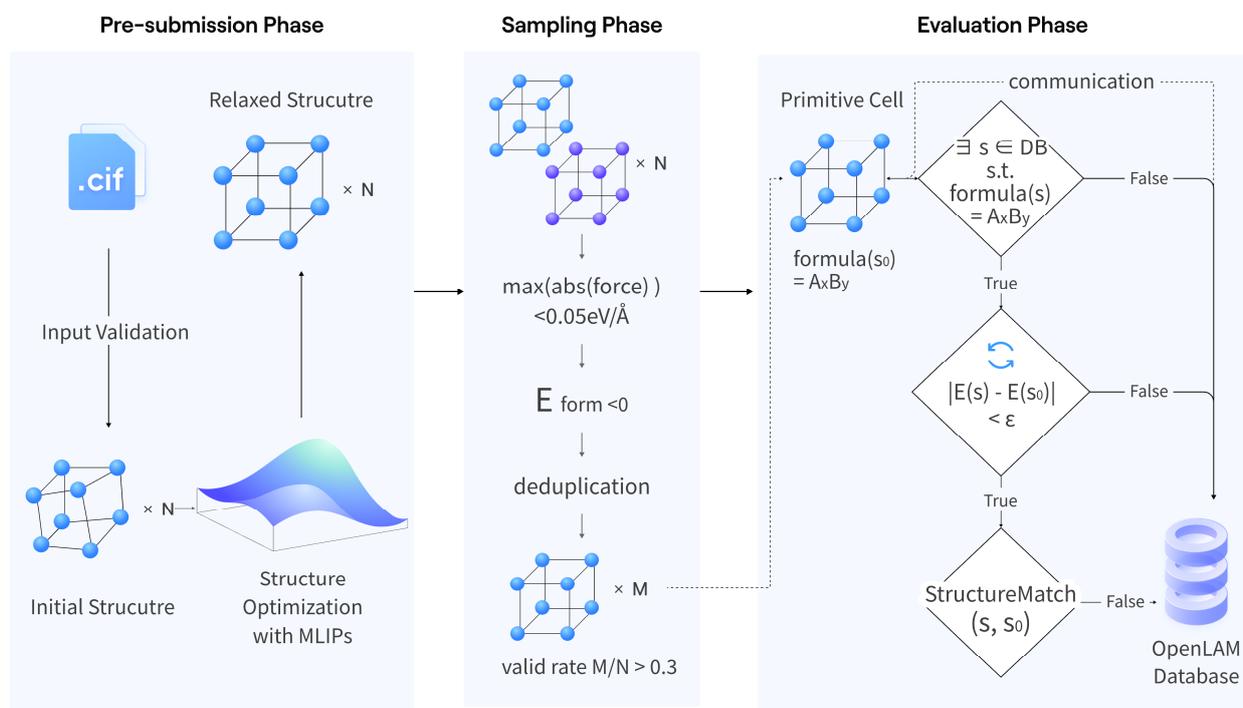

Figure 1. Schematic illustration of the three-phase validation and evaluation pipeline for crystal structure submissions in LAM Crystal Philately: pre-submission validation, random sampling quality check, and final evaluation with MLIP-based energy calculations.

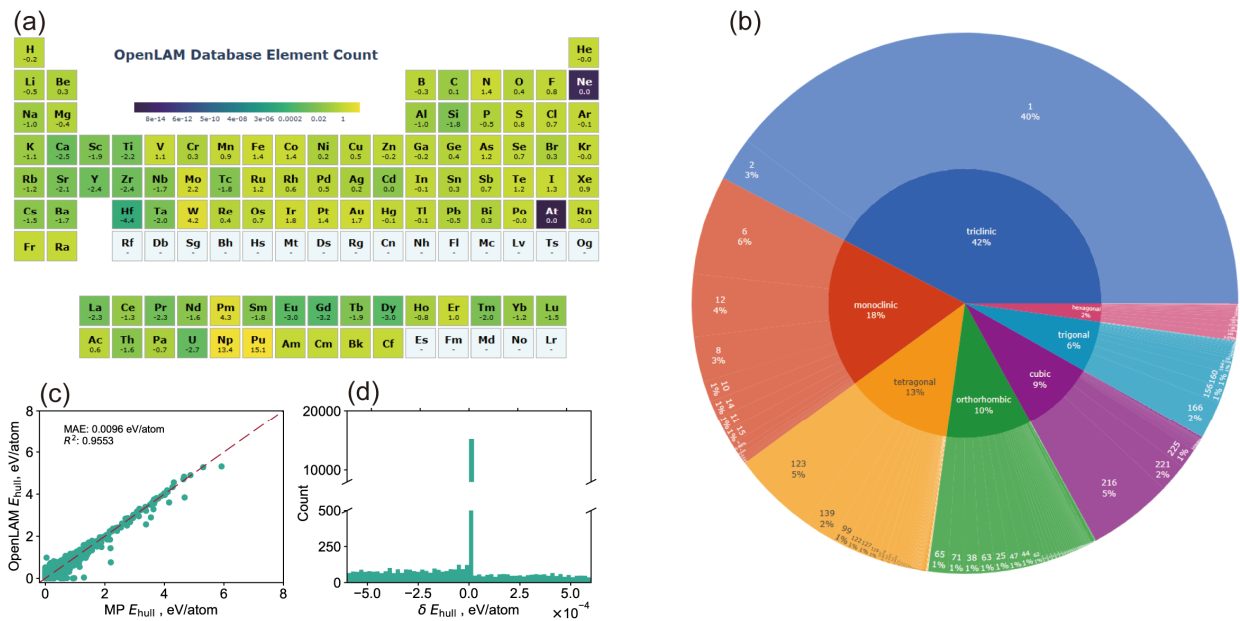

Figure 2. Statistical analysis of structures collected from the first round of LAM Crystal Philately competition. (a) Distribution of chemical elements and (b) crystallographic space groups across all valid submissions. (c) Correlation between the energy above the hull values from OpenLAM and the Materials Project (MP). (d) Distribution of energy deviations relative to the convex hull. Visualization: pymatviz. [14]

The first round of the LAM Crystal Philately competition has been successfully concluded, and the second round is scheduled to begin in February 2025. During the first round, we collected over 19.8 million valid structures, spanning a broad chemical and configurational space, (Figure 2a,b) among which 12.8 million are uploaded by the participants, while the rest were sourced from open repositories [15-30]. Our energy above hull predictions demonstrated strong consistency with the Materials Project (MP) database for known structures, validating the reliability of our MLIP-based evaluation framework. Given the reliability of our evaluation method, we can assert that, approximate 1.3 million participants-uploaded structures potentially expand the frontier of stable materials beyond what is currently documented in open-source structure repositories. (Figure 2c,d) To access and make use of those data, researchers can refer to the following resources. For accessing the data through the user interface (UI), the Crystal Craft APP [31] was developed, allowing users to query the collected structures via Wyckoff positions, chemical formulas, sources, or elements. As for accessing the data via application programming interface (API), the OpenLAM repository offers the essential interface [8].

Participants leveraged diverse generative algorithms to target specific objectives, such as creating ionic crystals with low formation energy when that was the scoring criterion, or designing high-entropy alloys when energy above the hull became the focus. Among the algorithms used, two notable ones are ConCDVAE[32] and InvDesFlow[33]. ConCDVAE has topped the leaderboard for several consecutive weeks, mainly due to its unique generation algorithm involving directionally generating structures with lower formation energy As for InvDesFlow, it was capable of generating structures with both lower formation energy and lower convex hull energy, thus gaining an edge in the competition. This process revealed an intriguing interplay between participant strategies and the competition rules, demonstrating the platform's dynamic and adaptive nature. We believe this evolving framework will serve as a valuable resource for researchers advancing generative models and foundational models in materials science.

## 3. Future Outlook

Material discovery is not a novel task for MLIPs, as several projects [34] have been developed for this purpose. However, a gap remains between the materials discovered and their practical applications. We aim to build upon this database to enable as many downstream applications as possible. Key directions that we are actively exploring and welcomes collaboration on, include: (1) Comprehensive Benchmark Construction: Developing robust benchmarks for evaluating material models; (2) Condition-Generation base on Experimental Properties: Tailoring ML models to generate conditions for experimental scientists; (3) Large-Scale Pretraining of Models (LAMs): Advancing the pretraining of Large Atom Models to enhance predictive capabilities across diverse material types.

We believe that these efforts will help to bridge the gap between material discovery and real-world applications.

**Acknowledgements**

We would like to express our special thanks to Dr. Lei Wang, Zhendong Cao, Jian Lv, Zhenyu Wang, Xiaoshan Luo, and Shigang Ou for their valuable suggestions. We are also grateful to Sikai Yao, Hui Zhou, and Zhaohan Ding for their contribution in developing infrastructures. The work of H.W. is supported by the National Key R&D Program of China (Grant No.~2022YFA1004300).